\pdfoutput=1

\documentclass[11pt]{article}

\usepackage[]{latex/acl}
\usepackage{times}
\usepackage{latexsym}
\usepackage{graphicx}
\usepackage[T1]{fontenc}

\usepackage[utf8]{inputenc}

\usepackage{microtype}

\usepackage{inconsolata}

\usepackage{amsfonts}       
\usepackage{subcaption}
\usepackage{multirow}
\usepackage{mathtools}
\usepackage{amsthm}
\usepackage{xspace}
\usepackage{enumitem}

\usepackage[acronym]{glossaries}

\glsdisablehyper

\usepackage{siunitx}
\usepackage{tikz}

\usepackage{booktabs}
\theoremstyle{plain}

\theoremstyle{definition}

\theoremstyle{remark}

\newcommand{\comment}[1][]{\textcolor{blue}}

\newcommand{\todo}[1][]{\textcolor{red}}

\newcommand{\orcid}[1]{\href{https://orcid.org/#1}{\textsuperscript{\includegraphics[height=2\fontcharht\font`A]{figures/orcidlogo.pdf}}}}

\newcommand{\Math}[1]{\ensuremath{#1}\xspace}

\newcommand{\ReLU}{\Math{\operatorname{ReLU}}}
\newcommand{\GELU}{\Math{\operatorname{GELU}}}
\newcommand{\Enc}{\Math{\operatorname{Enc}}}
\newcommand{\Dec}{\Math{\operatorname{Dec}}}

\newcommand{\softmax}{\Math{\operatorname{Softmax}}}
\newcommand{\LayerNorm}{\Math{\operatorname{LayerNorm}}}
\newcommand{\BatchNorm}{\Math{\operatorname{BatchNorm}}}

\newcommand{\R}{\Math{\mathcal{R}}}

\newacronym{CE}{CE}{Cross-Entropy}
\newacronym{CXR}{CXR}{chest X-Ray}
\newacronym{DL}{DL}{Deep Learning}
\newacronym{CNN}{CNN}{Convolutional Neural Network}
\newacronym{GC}{GC}{Garbled Circuits}
\newacronym{HE}{HE}{Homomorphic Encryption}
\newacronym{FHE}{FHE}{Fully Homomorphic Encryption}
\newacronym{KD}{KD}{Knowledge Distillation}
\newacronym{ML}{ML}{Machine Learning}
\newacronym{MPC}{MPC}{Multi-Party Computation}
\newacronym{NAS}{NAS}{Network Architecture Search}
\newacronym{SC}{SC}{Skip-Connection}
\newacronym{OT}{OT}{Oblivious Transfer}
\newacronym{PPML}{PPML}{Privacy-Preserving Machine Learning}
\newacronym{ResNet}{ResNet}{Residual Net}
\newacronym{SIMD}{SIMD}{Single Instruction Multiple Data}
\newacronym{SS}{SS}{Shared Secret}
\newacronym{ZSL}{ZSL}{Zero-Shot Learning}
\newacronym{SOTA}{SOTA}{State-of-the-Art}
\newacronym{SP}{SP}{Service Provider}
\newacronym{FFN}{FFN}{Feed-Forward Network}

\title{Converting Transformers to Polynomial Form for Secure Inference Over Homomorphic Encryption}

\author{
  Itamar Zimerman \\
  IBM Research\\
  Tel-Aviv University \\
  \And
  Moran Baruch \\
  IBM Research\\
  Bar-Ilan University \\
  \And
  Nir Drucker \\
  IBM Research\\
  \AND
  Gilad Ezov \\
  IBM Research\\
  \And 
  Omri Soceanu \\
  IBM Research\\
  \And
  Lior Wolf \\
  Tel-Aviv University \\
}
\begin{document}
\maketitle
\begin{abstract}
Designing privacy-preserving deep learning models is a major challenge within the deep learning community. \gls{HE} has emerged as one of the most promising approaches in this realm, enabling the decoupling of knowledge between the model owner and the data owner. Despite extensive research and application of this technology, primarily in convolutional neural networks, incorporating HE into transformer models has been challenging because of the difficulties in converting these models into a polynomial form. We break new ground by 
introducing the first polynomial transformer, providing the first demonstration of secure inference over HE with transformers. This includes a transformer architecture tailored for \gls{HE}, alongside a novel method for converting operators to their polynomial equivalent. This innovation enables us to perform secure inference on LMs with WikiText-103. It also allows us to perform image classification with CIFAR-100 and Tiny-ImageNet. Our models yield results comparable to traditional methods, bridging the performance gap with transformers of similar scale and underscoring the viability of HE for state-of-the-art applications. Finally, we assess the stability of our models and conduct a series of ablations to quantify the contribution of each model component. 
\end{abstract}

\section{Introduction}
\gls{PPML} has become a crucial field, ensuring that valuable insights can be inferred from data without compromising individual privacy. \gls{HE} stands out within this domain, offering the capability to compute on encrypted data, thereby safeguarding sensitive information during analysis. Modern HE schemes such as CKKS~\cite{ckks2017} support computation over encrypted input only when the computations are represented by polynomial functions. This limitation poses a unique challenge for \gls{DL} applications. For example, the non-linear activations such as GELU and Softmax are also non-polynomial and, therefore, must be adapted into an equivalent polynomial form. This adaptation is essential before employing them with encrypted data in a \gls{HE} framework. Understanding these theoretical limitations is important when we consider the practical applications of \gls{HE} in real-world scenarios, such as in NLP.

In the application of \gls{HE} within NLP, the model owner provides a trained DL model. The data owner, who wishes to use this model, encrypts their input data and sends it to the model owner for inference. The model owner processes this encrypted data. Neither the input nor the output is exposed to the model owner, who sends back the encrypted result to the data owner, who can then decrypt and interpret it.
\begin{figure}[h]
    \centering
    \includegraphics[width=0.95\linewidth]{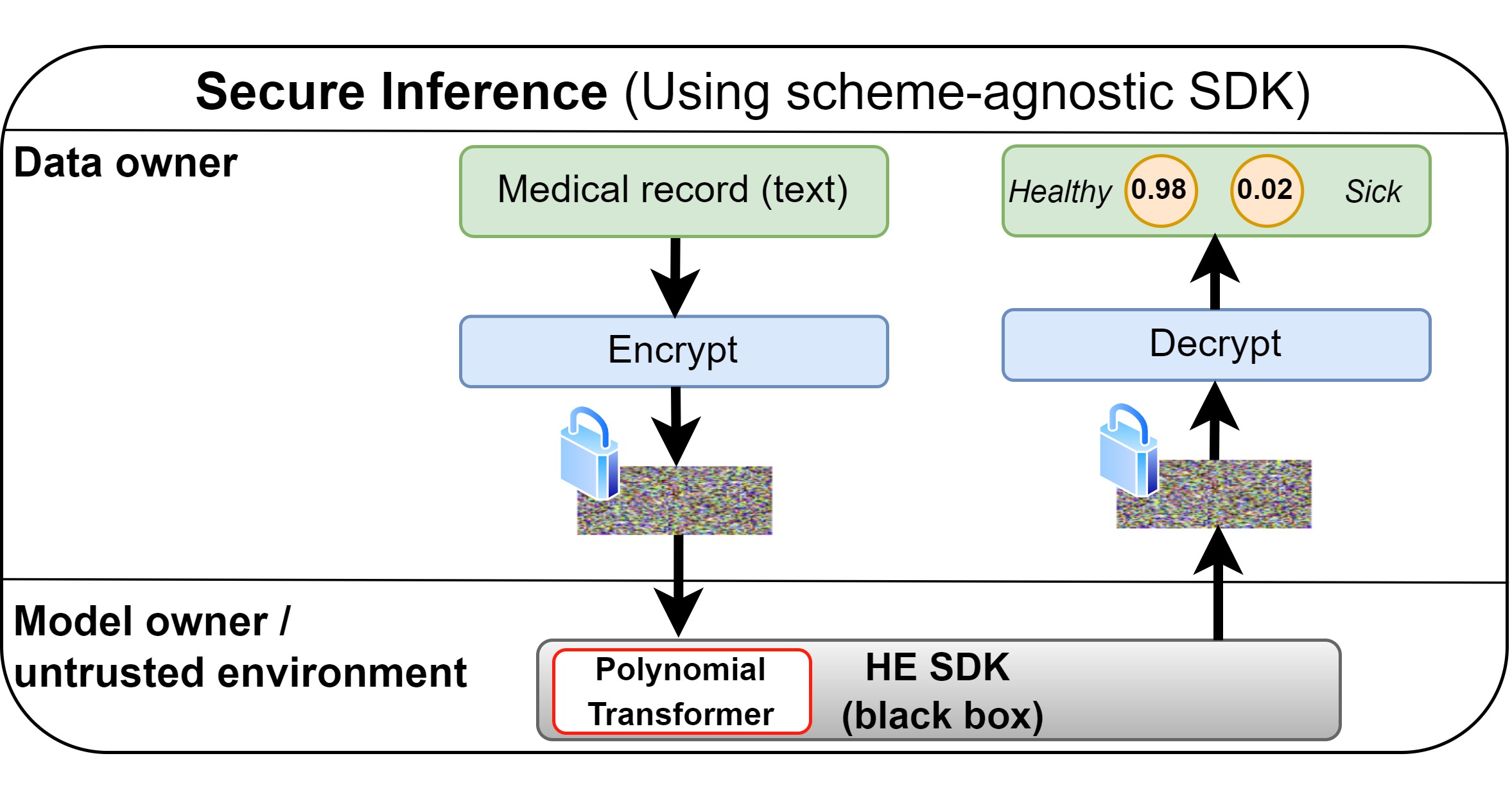}
    \caption{\textbf{Schematic Overview of Secure Inference Using Fully Homomorphic Encryption (FHE)}: The diagram depicts the sequence of steps where the data owner sends an encrypted sensitive data sample, as ciphertext, to the model owner or an untrusted environment. Within this environment, the HE-SDK employs the Polynomial Transformer to process the encrypted data, ensuring no access to the sensitive information. After computation, the encrypted result, obtained from the Polynomial Transformer, is returned to the data owner, who decrypts it to retrieve the final classification outcomes. As such, the privacy of the underlying data is maintained throughout the entire process.}
    \label{fig:threatModel}
\end{figure}

As of now, the adaptation process to align with the limitations of \gls{HE} has been primarily focused on polynomial CNN models such as AlexNet~\cite{alexnet}, ResNet~\cite{resnet}, and ConvNext~\cite{convnext1}. This focus is evident in various previous works, e.g.~\cite{baruch2023sensitive,helayers,pmlr-v162-lee22e,baruch2021fighting,chet_compiler,CryptoNets2016}. These models commonly use \ReLU and \GELU activations, which are relatively straightforward for polynomial approximation. However, transformer-based models~\cite{vaswani2017attention} remain a notable exception. Due to their inherent non-polynomial operations such as Softmax -- which, unlike \ReLU, requires division by an exponent and presents a more challenging case for polynomial approximation -- and their large structural size, these models pose substantial challenges in HE adaptation. For example, one study ~\cite{zhou2019polynomial} showed that polynomial models tend to be less stable with increasing depth and width.


{\noindent \textbf{Our contributions.}\quad}We introduce a novel approach to adapt transformers to be compatible with modern \gls{HE} schemes for secure inference. Through a series of simplifications and approximations, we introduce methods to create, for the first time, practical polynomial transformer models that retain competitive performance across both language modeling and image classification. We demonstrate the feasibility of employing transformers under \gls{HE} while bridging the performance gap with their non-encrypted counterparts of similar scale. By providing stability analyses and ablations, we provide a comprehensive understanding of the trade-offs and potentials of using polynomial transformers in privacy-preserving settings, charting a course for future innovations in \gls{PPML}. Finally, we provide techniques for polynomial and HE-friendly alternatives to the layer normalization and self-attention layers. Such techniques can enhance existing cryptographic protocols and improve client-aided solutions or may facilitate the development of a wider variety of polynomial models, extending beyond transformers.
\section{Background}
\subsection{\acrfull{FHE}}
\paragraph{\acrfull{HE}.} We start by describing the high-level background and basic concepts of \gls{HE} schemes.
\gls{HE} schemes allow us to perform operations on encrypted data~\cite{Gentry2009}.
The \gls{HE} system has an encryption operation $\Enc:\R_1 \rightarrow \R_2$ that encrypts plaintext input from the ring $\R_1(+, *)$ into ciphertexts in the ring $\R_2(\oplus, \odot)$ and an associated decryption operation $\Dec:\R_2 \rightarrow \R_1$. An \gls{HE} scheme is correct if for every valid input $x,y \in \R_1$

\begin{align}
& \Dec(\Enc(x)) = x\\
\label{eq:add}& \Dec(\Enc(x) \oplus \Enc(y)) = x + y \\
\label{eq:mul}& \Dec(\Enc(x) \odot \Enc(y))  = x * y\,
\end{align}
and is approximately correct if for some small $\epsilon > 0$ that is determined by the key, it follows that $|x - \Dec(\Enc(x))| \le \epsilon$, and similarly modifying Eqs.~\ref{eq:add}, and \ref{eq:mul}. We used the CKKS scheme \cite{ckks2017} in this paper's experiments.
Note that each multiplication operation in \gls{HE} ($\odot$) adds noise to the ciphertext, potentially impacting decryption accuracy if not managed. 
\textit{\gls{FHE}} supports the execution of a theoretically unlimited number of both addition and multiplication operations on encrypted data, without the need for decrypting intermediate results. This is achieved through a computationally intensive process known as \textit{bootstrapping}, which refreshes the ciphertext to prevent noise buildup and maintain decryption accuracy. Thus, \gls{FHE} can evaluate complex arithmetic circuits on encrypted data, making it a powerful tool for secure, privacy-preserving computation.

\subsection{Polynomial DL Models}
Producing polynomial networks with high accuracy is challenging, and several theoretical intuitions and proofs were proposed. For example, \citet{zhou2019polynomial} proved that under some conditions polynomial FFNs are unstable, concluding that the likelihood of instability in a polynomial network increases with its complexity, specifically as depth and width grow.
\citet{goyal2020improved} suggested that the problem with poly-activations is that the gradients and outputs are unbounded and can be arbitrarily large, unlike other activations such as ReLU and GELU. 
They also pointed out that in deeper networks $f_{(d,l)}$ with $l$ layers and $d$-degree polynomial activations, the gradients explode exponentially in the degree of the entire network, since for input $ x > 1, \lim_{x \rightarrow \infty} f_{(n.l)}(x)/x=\infty$. Additionally, \citet{chrysos2020p, goyal2020improved, gottemukkula2020polynomial} attempted to implement deep polynomial networks but faced optimization instability. They resolved the issue by incorporating non-polynomial components into their models.

Recent works focus on converting Deep CNNs into polynomials. The method in \cite{baruch2023sensitive} introduced a technique to stabilize polynomial models by adding a loss term that minimizes the input range to the non-polynomial layers. Using this approach, they successfully produced low-degree polynomial versions of ResNet-152 and ConvNeXt on ImageNet. In addition, \cite{lee2021precise} approximated \ReLU using a composition of three polynomials to precisely approximate \ReLU, achieving high-degree polynomial models. To the best of our knowledge, these works represent some of the deepest polynomial models to date. However, none of these or any other works have tackled the problem of polynomial transformers.

\subsection{Polynomial Approximation}
Polynomial networks are commonly obtained by approximating the non-polynomial functions of pre-trained networks, e.g., \cite{relu1, relu2, SecureML, cryptoDL,lee2021precise}, or by substituting ReLU during or after a dedicated training process ~\cite{baruch2021fighting,baruch2023sensitive}. The Remez algorithm \cite{Remez, RI1, RI2} is commonly used for finding the optimal polynomial approximation of a function $f(x)$ in a certain degree within a pre-defined range $[a,b]$, assuming a uniform distribution of $x$. Alternatively, iterative methods such as the Newton–Raphson method \cite{raphson1702analysis} offer another polynomial approximation approach. Specifically, \cite{panda2022polynomial} focused on approximating $\frac{1}{\sqrt{x}}$ in the interval $[a,b]$, by dividing the interval into sub-intervals and approximating over each via Newton–Raphson method, before aggregating the results with another polynomial. However, the input range to the non-polynomial layers $[a,b]$ can be extremely large, which results in poor and no practical approximations. 
This paper employs polynomials for \ReLU, as defined in \cite{lee2021precise}, for layer normalization (inverse square root) from \cite{panda2022polynomial} and for \GELU using polynomials derived from the Remez algorithm.

\subsection{Transformers in PPML}
The integration of transformers into \gls{PPML} solutions has become significantly prominent, highlighting the relevance of both fields. In recent years, a variety of secure interactive protocols have been introduced to enable secure inference of transformer models ~\cite{TransformerPPML1, TransformerPPML3, TransformerPPML4, TransformerPPML5, TransformerPPML6, TransformerPPML7, TransformerPPML8, TransformerPPML9}. These methods leverage mechanisms such as shared secrets to compute non-polynomial operations like \GELU, \softmax, and layer normalization. However, such approaches increase communication overhead and the potential for vulnerability to man-in-the-middle attacks. Our method addresses these concerns by enabling computation in untrusted environments, eliminating the need for additional communication, and thus preserving a non-interactive stance. By offering polynomial alternatives for non-polynomial operations in transformers, our method not only enhances existing protocols, but also eliminates the need for client involvement in the computation process of non-polynomial operations. One alternative approach involves applying secure inference on text embeddings extracted from an unsecured BERT transformer via a simple HE-based classification model~\cite{lee2022privacy,lee2023hetal}. However, this method addresses a significantly narrower threat model.

\section{Problem Settings\label{sec:problem}}
We begin by clearly defining the motivation and problem settings before discussing our methodology. Our objective is to develop a transformer model that uses only polynomial operations and performs well on downstream tasks. By utilizing these polynomial-based transformers, we aim to enable secure inference within the \gls{HE} framework. Note that this paper does not cover secure training. One might think that replacing non-polynomial operations in the network with polynomial alternatives or approximated polynomials, either before or after training, could simply solve the problem. However, polynomial networks are unstable, and instability issues can arise during training or when non-polynomial operations are replaced, especially in deep networks (as analyzed from both theoretical and empirical perspectives in~\cite{zhou2019polynomial,goyal2020improved}). Therefore, standard methods for creating polynomial deep learning models for secure inference involve several architectural modifications and unique training procedures, including training of non-polynomial networks as an intermediate step~\cite{baruch2023sensitive, baruch2021fighting, ao2023autofhe}. Furthermore, in general, as higher-degree polynomials are used, the running time for secure inference increases drastically. Hence, a common challenge is to reduce both the polynomial degree of each operation and the overall multiplication depth of the model.

\section{Method\label{sec:method}}
Addressing the problem defined in Section~\ref{sec:problem}, our methodology begins by identifying non-polynomial components in the transformer model: (i) The \softmax function in the attention is non-polynomial, involving exponentiation and division; (ii) layer normalization (\LayerNorm), which normalizes features by dividing them by their standard deviation, also contains the square root function, adding to the non-polynomial complexity; and (iii) activation functions, which are traditionally non-polynomial but have been substituted with polynomial alternatives in prior research~\cite{lee2021precise,baruch2023sensitive,lee2022low}.

From earlier stages of our research, we found that directly approximating the \softmax and \LayerNorm by polynomials within each transformer block is challenging due to: (a) the inherent complexity of these functions, which involve multiple non-polynomial operations—specifically, division and exponentiation for \softmax, and division along with square root for \LayerNorm, and (b) the nature of these functions as vector-based rather than scalar-based, unlike neural activation functions, which were approximated by polynomials previously. In light of these complexities, our work seeks to develop HE-friendly alternatives to \softmax-based attention (in Section~\ref{subsec:HEFriendlyAttention}) and \LayerNorm (in Section~\ref{subsec:heFrindelyNormalization}) that are easier to approximate by a polynomial, ideally employing operations that are polynomial in nature, as well as those that can be precisely approximated by polynomials. Our entire pipeline for polynomial adaptation is detailed in Section~\ref{subsec:FullPipeline}.

\begin{figure*}[ht]
    \centering
    \includegraphics[width=1.0\linewidth]{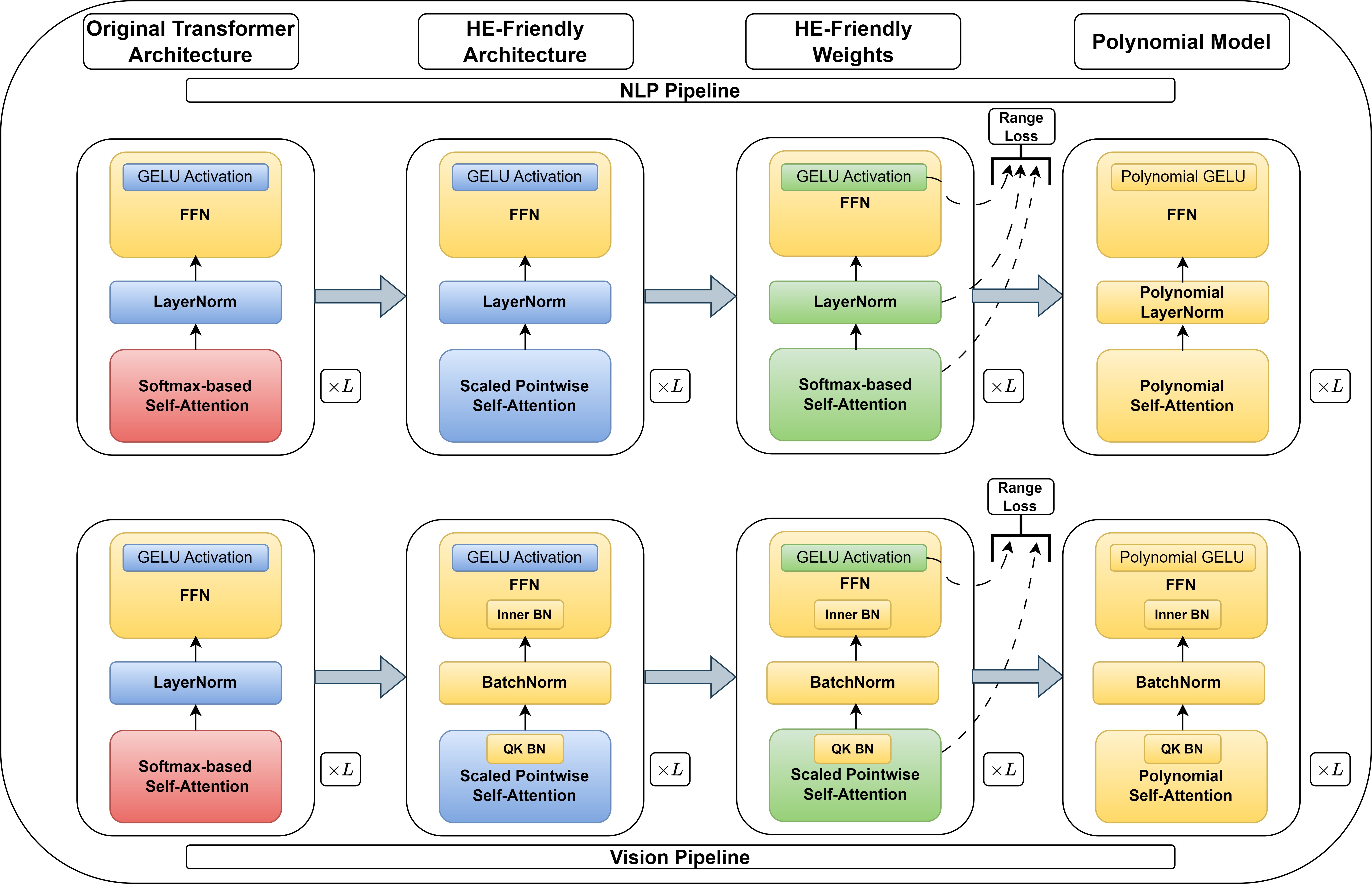}
    \caption{\textbf{Method:} The NLP pipeline is depicted at the top, and the vision pipeline shown at the button. \textbf{Yellow} marks  polynomial components, while \textbf{red} indicates components that cannot be efficiently adapted to polynomial form, and therefore replaced by components that can be replaced by a polynomial via range-minimization tuning (\textbf{blue}). \textbf{Green} components can be easily replaced by polynomials, given their current weights. Only the two central columns include training, while the rest include component replacement. Skip-connections are omitted for clarity.}
    \label{fig:methodFig}
\end{figure*}

\subsection{HE-Friendly Attention\label{subsec:HEFriendlyAttention}}
To circumvent the use of \softmax attention, we employ a pointwise activation-based attention, denoted as $\text{Attention}_{\sigma}$ as an alternative. This \softmax-free mechanism can be formalized as:
\begin{equation}
\text{Attention}_{\sigma}(Q, K, V) = \sigma\left(\frac{QK^T}{\sqrt{d_k}}\right)V
\end{equation}
where \(\sigma\) acts as an element-wise activation function, differing from the vector-wise \softmax function used in standard attention mechanisms. This modification simplifies the transformation into a polynomial form, effectively reducing the problem from polynomial attention to the well-studied problem of handling polynomial activation functions. This mechanism enhances model accuracy~\cite{ma2022mega} and reduces latency~\cite{hua2022transformer,so2021primer}, which are benefits outside the scope of HE. However, those implementations have combined it with additional techniques, such as gated attention and extra unique normalization. These additions result in an increase in the overall multiplication depth and require the development of additional HE-friendly alternatives. In our work, we omit these supplementary components altogether. In their absence, the standalone mechanism becomes less stable, and we use length scaling to overcome this instability:

\begin{equation}
\text{Attention}_{\sigma\text{-scale}}(Q, K, V) = \frac{1}{\text{S}(L)} \sigma\left(\frac{QK^T}{\sqrt{d_k}}\right)V\,,
\end{equation}
where \( L \) represents the sequence length, and \(S \) is a scaling function defined as either \( \frac{1}{\sqrt{L}} \) or \( \frac{1}{L} \). Thus, the stabilizing effect of the \softmax function is replaced by a scaling factor that compensates for the instability introduced by the alternative attention mechanism. Additionally, as the multiplication of the attention matrix with the values matrix involves summation over \( L \) elements, it is logical to normalize by this factor.
We investigate various scaling functions, applied both pre- and post-activation, and can be implemented by modifying the $\sigma$ activation function:
\begin{equation}
\text{Pre-act scaling: }  \hat{\sigma}(x) = \sigma(S(L)x)
\end{equation}
\begin{equation}
\text{Post-act scaling: }  \hat{\sigma}(x) = S(L)\sigma(x)
\end{equation}
\begin{equation}
\text{Pre- and post-scaling: }  \hat{\sigma}(x) =  S(L)\sigma(S(L)x)
\end{equation}

The selection of scaling functions is determined by empirical considerations and stability analysis

\paragraph{Choosing the Attention Activation}
Naturally, we initially experimented with polynomials as activations. We started with polynomials used in previous HE-literature, ranging from a simple quadratic activation ($x^2$) ~\cite{CryptoNets2016} to high-degree polynomials~\cite{lee2021precise} that approximate \ReLU and other standard activations. However, these polynomial activations were found to be unstable during training. To address this, we first trained our model using standard non-polynomial activations and then applied an additional training phase to convert these activations into polynomials, similar to~\cite{baruch2023sensitive}. This approach balanced performance in the initial training phase with the precision of polynomial approximation in the later phase.

\paragraph{Reformulate Attention Mask\label{par:ReformulateAttentionMask}}
Traditional practices, such as those employed in the Swin transformer or in training LLMs via self-supervised learning, manipulate self-attention via masking to determine which tokens can attend to each other. These standard mask mechanisms are specifically designed for the Softmax-based self-attention and should be reformulated for pointwise attention:
\begin{equation}
\text{Attention}_{\sigma}(Q, K, V) = \left(\sigma\left(\frac{QK^T}{\sqrt{d_k}}\right) \odot M \right) V
\end{equation}

where $M$ is the binary mask. This mechanism is agnostic to any type of pointwise activation $\sigma$.

\subsection{HE-Friendly Normalization\label{subsec:heFrindelyNormalization}}
To enhance training stability, transformers rely on \LayerNorm, which is formulated as follows:
$$
\text{\LayerNorm}(x) = \frac{x - \mu}{\sqrt{\sigma^2 }} \cdot \gamma + \beta
$$
where $x$ is the input vector, $\mu$ is the mean of $x$, $\sigma^2$ is the variance, and $\gamma$ and $\beta $ are learnable parameters. 
Computing \LayerNorm over HE requires calculating the inverse square root, which is not a polynomial operation. A common practice in designing neural networks for secure inference over HE is to replace \LayerNorm with \BatchNorm, as it can be implemented by a straightforward constant affine transformation at inference time. Therefore, we attempted to train transformers using $\sigma$-attention and \BatchNorm. We observed that these models were highly unstable, performing poorly on vision tasks and failing to converge in NLP tasks. Consequently, we adopt two distinct approaches for vision and NLP tasks.

\paragraph{Normalization for Vision Transformers}
For vision transformers, to improve performance and mitigate training instability, we add two components: (i) Additional \BatchNorm in the MLP of ViT, which is proposed in~\cite{yao2021leveraging} as a stabilizer for ViT training, and (ii) additional \BatchNorm within the $\sigma$-attention, which normalizes values across different attention heads, since we observe that those are the sources of instability. The resulting $\sigma$-attention variant is:
\begin{equation}
\frac{1}{\text{S}(L)} \sigma\left(\text{\BatchNorm2D}\left(\frac{QK^T}{\sqrt{d_k}}\right)\right) V
\end{equation}

\paragraph{Normalization for NLP Transformers}
For NLP, models with $\sigma$-attention and \BatchNorm completely failed, even when augmented by stabilizing factors from the literature, such as~\cite{wang2022understanding}. Consequently, we had to confront the challenge of approximating \LayerNorm by polynomials, which entails approximating the inverse square root function. Empirically, we found that the values of the variance in trained transformers (with $\sigma$-attention) ranged between $1$ and $10^9$, causing approximation challenges due to the extremely large domain. To solve this problem, we first focus on narrowing the domain of the variance, which then makes it easier to approximate the inverse square root over this restricted domain. The method is similar to~\cite{baruch2023sensitive}, which introduces an additional loss function that encourages the model to minimize the range of the input to activation layers. We apply this technique on the variance at each layer via the following objective: 
\begin{equation}
\label{eq:LNLoss}
\mathbb{L}_{\text{Variance Minimization}} := \Sigma_{m=1}^{L_N} \max_{c \in C, x_i \in X} \left( {\text{var}^{i}_{m,c}}\right)
\end{equation}
where we denote the number of layers by $N_L$, the number of channels by $C$, and the train dataset by $X := [x_1, x_2, ..]$. Furthermore, we denote the variance at layer $m$ and channel $c$, when the model processes the $x_i$ example by $\text{var}_{m,c}^{i}$. For reasons of efficiency, we compute the loss over each batch rather than the whole training set $X$. By extending this method to operate on layer normalization instead of activations, we succeed in reducing the variance range to a smaller domain. 
This reduction makes it feasible to use well-known approximations, such as the technique described in~\cite{baruch2023sensitive}, for the inverse square root.

\subsection{A Recipe for a Polynomial Transformer\label{subsec:FullPipeline}}
Fig.~\ref{fig:methodFig} illustrates the entire method, which comprises three stages: (i) First we modify the architecture from the original transformer architecture (first column) to a \textbf{HE-friendly architecture} (second column), namely, an architecture that can eventually be converted into a polynomial form. Then we train the modified model from scratch with the same hyperparameters. (ii) In the second stage, we perform a supplementary training procedure to obtain a model with \textbf{HE-friendly weights}, which means that each non-polynomial component will only operate on specific and restricted domains. To do so, we add a loss function that minimizes the range of inputs to non-polynomial layers. For the activations (standard activations and attention-activations), we directly apply the method from~\cite{baruch2023sensitive}, which defines the range loss for activations. For the \LayerNorm layers, we use the loss defined in Eq.~\ref{eq:LNLoss}. The whole training objective $\mathbb{L}$ is defined by: $$ \alpha \mathbb{L}_{\text{Range Minimization}} +  \beta \mathbb{L}_{\text{Variance Minimization}} + \mathbb{L}_{\text{original}}$$ where $\alpha$ and $\beta$ are hyperparameters. 
(iii) Finally, each non-polynomial layer is directly replaced with its polynomial approximation, resulting in a \textbf{polynomial model}. Appendix~\ref{sec:appendixPolynomials} contains details on the approximation we used. Those approximations are accurate for the HE-friendly architecture \& weights obtained from earlier stages.  

\section{Experiments}
We evaluate the polynomial models generated by our method in Section~\ref{subsec:resPolyModels}, focusing on language modeling with the Wikitext-103 dataset and image classification using standard benchmarks, including Tiny-ImageNet and CIFAR-10. Section~\ref{subsec:ResModelAnalysis} justifies our methodological choices, specifically the use of scaled $\sigma$-attention and an additional training phase designed to manipulate the input values of non-polynomial layers. Furthermore, that section contains several ablation studies to assess the impact of each method component on the overall performance degradation. Section~\ref{subsec:resultHE} discusses the accuracy and latency implications of applying our models over \gls{FHE}. The experimental setup is detailed in Appendix~\ref{sec:experimentalSetup}.

\subsection{Polynomial Models\label{subsec:resPolyModels}}
{\noindent \textbf{Polynomial Language Modeling} \quad}
We evaluated our BERT-like transformer model for language modeling as our NLP task. Specifically, we trained on Wikitext-103 with a self-supervised scheme for Next Token Prediction (NTP). The results in Table~\ref{tab:resNLPTransfromer} show that after architectural and training modifications, we achieved a fully polynomial model with competitive perplexity scores. In particular, the perplexity increased by 0.91 compared to a vanilla transformer of the same size, from 18.98 to 19.89 for a 6-layer transformer (53.3M parameters), and by 2.02 from 16.89 to 18.91 for a 12-layer model (95.8 Mparameters). Considering that at least 80\% of the gap between the vanilla transformer and our corresponding polynomial model is caused in the last stage where polynomial approximations are used (0.74 for 6 layers model and 1.76 for 12 layers), we hypothesize that more accurate polynomials can mitigate most of the performance gap. 
\begin{table}[h]
\centering
\begin{tabular}{ccccc}\toprule
\textbf{Depth} & \textbf{Original} & \textbf{P} & \textbf{P+MR} & \textbf{Poly}\\
\midrule 
\midrule 
6 & 18.98 & 19.07 & 19.15 & 19.89 \\ 
12 & 16.89 & 16.98 & 17.15 & 18.91 \\ \bottomrule
\end{tabular}
\begin{tabular}{lc}
\hline
\end{tabular}
\caption{\textbf{NLP Results:} Perplexity results of a polynomial BERT-like transformer on the Wikitext-103 benchmark. `Depth' indicates the number of transformer layers. `Original' denotes the perplexity of the vanilla Softmax-based transformer of equivalent size. `P' represents models utilizing scaled $\sigma$-attention, while `P+MR' shows perplexity at the end of the range minimization training. `Poly' details the final performance after substituting \LayerNorm and activation functions with polynomial approximations.}
\label{tab:resNLPTransfromer}
\end{table}

{\noindent \textbf{Polynomial Image Classification} \quad}
We evaluated our vision models on two image classification benchmarks: Tiny-ImageNet and CIFAR-100. The results, presented in Table~\ref{tab:resPolyViT}, indicate that our vision models, which are converted to polynomial form by our methods, remain competitive. Specifically, for ViT on CIFAR-100, the original ViT (denoted as `O') achieved a score of 73.4\%, whereas our HE-friendly alternative (P+BN+QK+A), achieved a score of 71.1\%. The HE-friendly alternative employs \BatchNorm as the normalization layer, includes additional stabilizers described in~\ref{subsec:heFrindelyNormalization} and is based on scaled-$\sigma$ attention. After applying our range-aware training procedure, the accuracy of our model decreased marginally by 0.1\% to 71.0\%, and it further decreased to 70.8\% after approximating non-polynomial components. For the Swin Transformer on Tiny-ImageNet, the original model achieved 59.4\%, and transitioning to the HE-friendly architecture resulted in a performance decrease of 0.3\% to 59.1\%. After employing range-aware training to obtain HE-friendly weights, the performance further degraded by 0.2\% to 58.9\%, while the final performance of the polynomial model remained the same. In conclusion, the performance gap between the polynomial models and the original architectures is less than 4\%, demonstrating the practicality of our methods in this domain.
\begin{table}[h]
\centering
\begin{tabular}{@{}l@{~}l@{~~}c@{~}c@{~}l@{~}r@{}}
\toprule
Model & Dataset & {O} & {P+BN+QK+A} & {MR}  & {Poly} \\ \midrule
\midrule 
ViT & CIFAR-100 & 73.4 & 71.1 & 71.0 & 70.8 \\

Swin & Tiny-ImgNet& 59.4 & 59.1 & 58.9 & 58.9 \\\bottomrule
\end{tabular} 
\begin{tabular}{lc}

\end{tabular}
\caption{\textbf{Vision Results:} Test accuracy results of a polynomial ViT. `O' represents the original vanilla model, `P+BN+QK+A' represents scaled-$\sigma$ attention-based ViT trained with \BatchNorm as the normalization function instead of \LayerNorm, and contains the additional stabilizers described in~\ref{subsec:heFrindelyNormalization}. `MR' refers to the accuracy at the end of the range minimization training, and `Poly' details the final performance after substituting polynomial approximations.}
\label{tab:resPolyViT}
\end{table}

\subsection{Model Analysis\label{subsec:ResModelAnalysis}}
\begin{figure*}[h]
    \centering
    \includegraphics[width=1.0\linewidth]{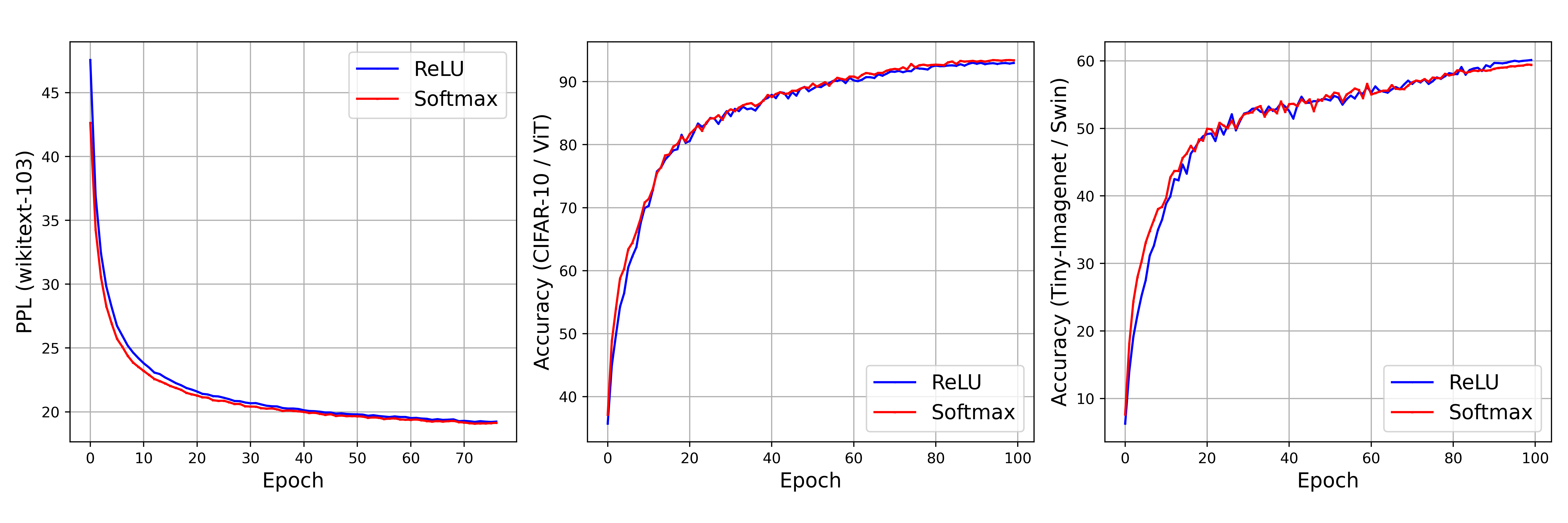}
    \caption{Scaled-$\sigma$ attention is comparable with Softmax attention. On each graph, we present the test accuracy for vision tasks, or text perplexity for language modeling tasks.}
    \label{fig:pointwizeVsSoftmax}
\end{figure*}

{\noindent \textbf{Scaled-$\sigma$ attention} \quad} We began our analysis by empirically assessing the performance differences between the vanilla transformer and the scaled-$\sigma$ attention. We conducted experiments in three regimes: (i) NLP, using the Wikitext-103 dataset with a 6-layer BERT-like transformer; (ii) image classification on the CIFAR-10 benchmark with the ViT backbone; and (iii) additional image classification tasks using the Tiny-ImageNet benchmark with a Swin-Transformer backbone. Across all regimes, we employed the same hyperparameters that were optimized for the vanilla transformer, which can be found in Table~\ref{tab:Visionhyperpams} and Table~\ref{tab:NLPhyperpams} (see Appendix~\ref{sec:experimentalSetup}). The training curves are presented in Figure~\ref{fig:pointwizeVsSoftmax}, and indicate that the models with scaled-$\sigma$ attention perform comparably to the baseline.

To justify our design choices regarding length scaling, we performed analyses in both the NLP and vision domains, comparing several variants of scaled-sigma attention models. These variants included models without length scaling and models with length scaling, employing the two functions $S(L) = \frac{1}{\sqrt(L)}$ and $S(L) = \frac{1}{L}$, applied at different positions relative to the activation function—either before, after, or both.

\begin{figure*}[h]
    \centering
    \includegraphics[width=0.7\linewidth]{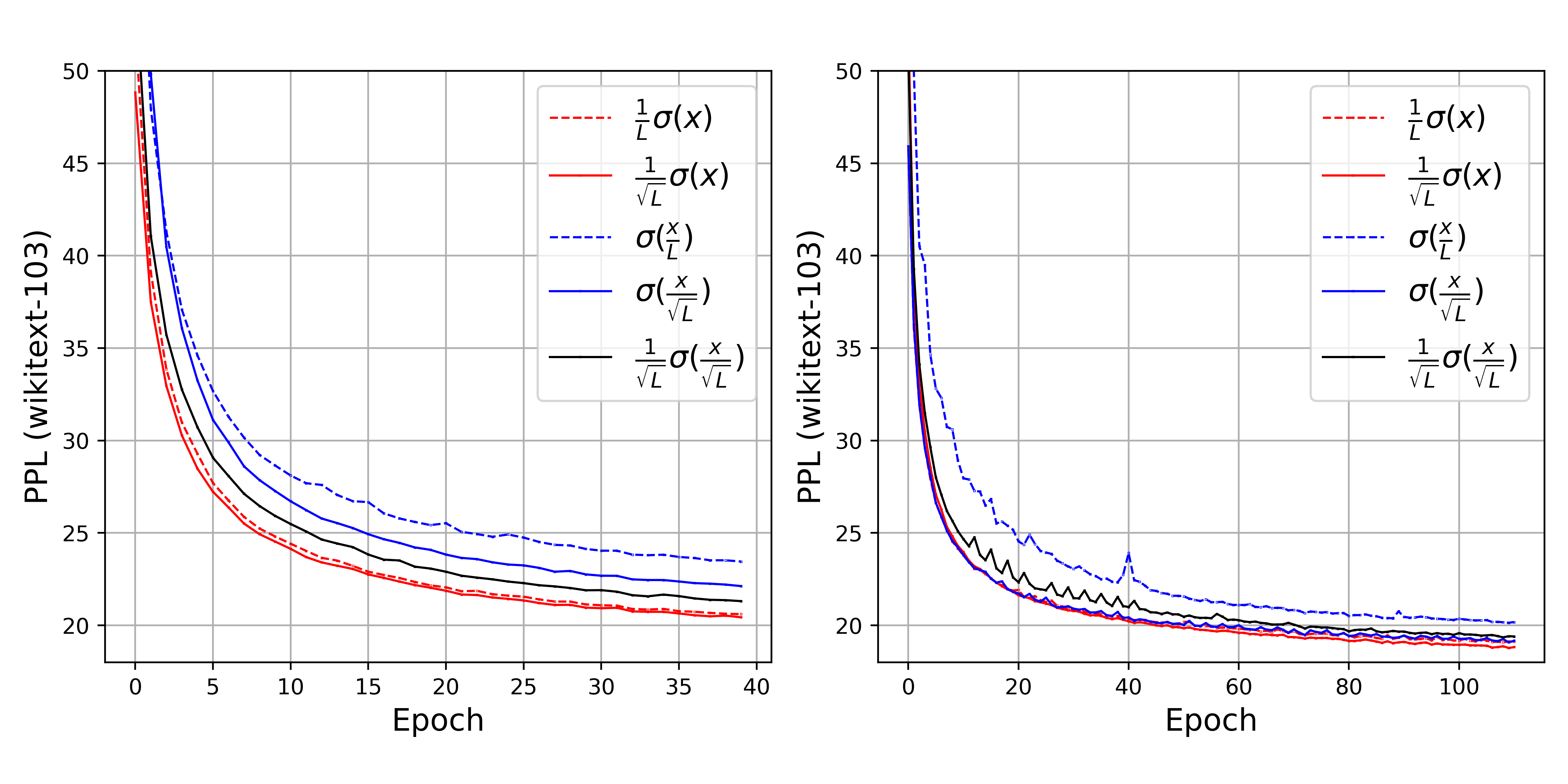}
    \caption{\textbf{Pointwise Transformers Require Scaling in NLP:}. Experiments with GELU attention are presented on the left and with Squared ReLU attention on the right. All experiments without scaling, or with both pre-scale and post-scale by length ($ \frac{1}{L} \sigma (\frac{x}{L}) $) collapse at an early stage in the training, even for lower learning rate.
    }
    \label{fig:PointwizeScaling}
\end{figure*}

For the NLP tasks, consistent with our previous experiments, we employed a 6-layer BERT-like transformer as a baseline and evaluated the variants on the Wikitext-103 dataset. We experimented with two types of activation functions: GELU and Squared ReLU. Notably, without length scaling, we observed that the model's weights explode in the initial epochs of training. This phenomenon confirms the necessity of integrating length scaling with $\sigma$-attention. Furthermore, variants utilizing both pre-scaling and post-scaling with a scaling function of $S(L)=\frac{1}{L}$ failed to converge, and eventually collapsed. For the other scaling functions, the results are depicted in Figure~\ref{fig:PointwizeScaling}. Evidently, employing post-length scaling with the scaling function $S(L)=\frac{1}{\sqrt{L}}$ provides the best performance for both types of attention activation.

For vision tasks, we replicated the settings detailed in Section~\ref{subsec:resPolyModels} with the ViT backbone. Test accuracy results with and without length scaling for attention activations are presented in Table~\ref{tab:ScalingInViT} for models evaluated on both the CIFAR-10 and CIFAR-100 benchmarks. These experiments were conducted with \BatchNorm, as this was the type of normalization we use for secure ViT (see Section~\ref{subsec:heFrindelyNormalization}). We employ post-activation length scaling with a scaling function $S(L)=\frac{1}{\sqrt{L}}$, which we found to be optimal for scaled-$\sigma$ based ViT. The findings underscore the significance of length scaling, which substantially enhances performance. Specifically, length scaling improves the GELU-attention models, increasing their accuracy by 2.46\% from 68.41\% to 70.87\% on CIFAR-100, and lifting their performance by 11.14\% from 81.17\% to 92.31\% on CIFAR-10.

\begin{table}[h]
\centering
\begin{tabular}{lccc}\toprule
 Datset      &  Vanilla    & GELU      & Scaled-GELU \\\midrule
\midrule
 CIFAR-100   &  72.08 & 68.41   & 70.87 \\ 
 CIFAR-10    &  92.70 & 81.17   & 92.31 \\ \bottomrule
\end{tabular}
\caption{\textbf{Pointwise ViT Require Scaling:} Experiments on both CIFAR-100 and CIFAR-10 with ViT. For both benchmarks, we compare vanilla attention, GELU-attention, and scaled-GELU attention.}
\label{tab:ScalingInViT}
\end{table}

\begin{figure*}[h]
    \centering
    \includegraphics[width=0.95\linewidth]{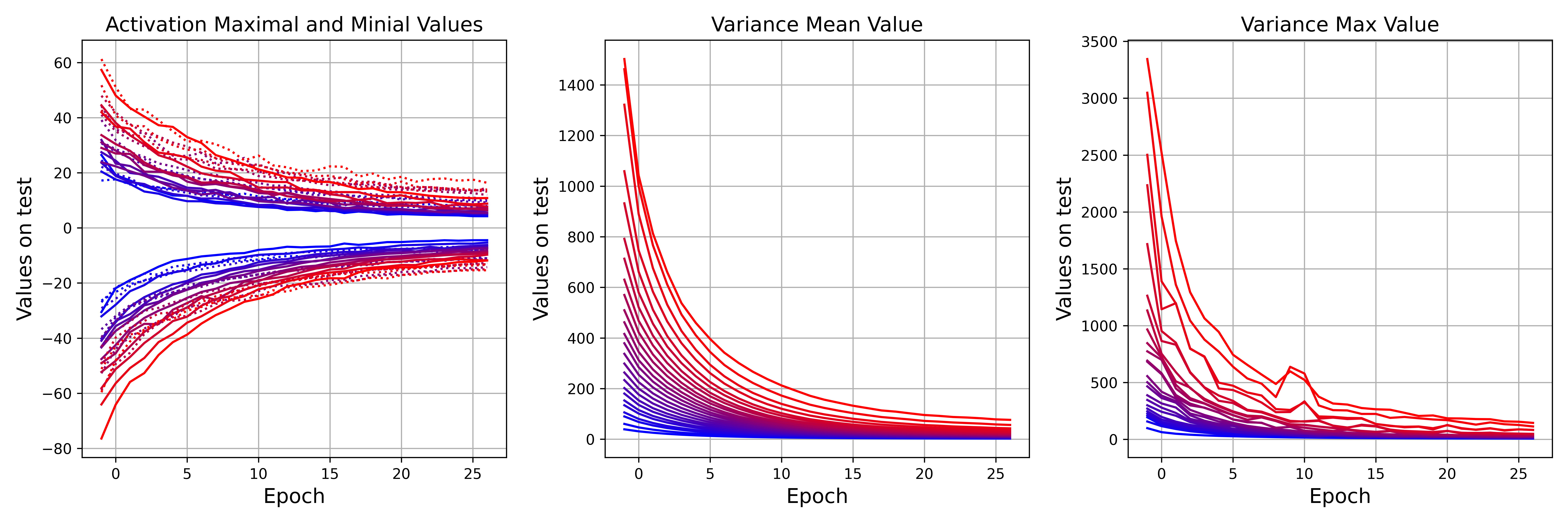}
    \caption{\textbf{Range Minimization:} The graph illustrates the impact of applying range regularization loss to constrain the input range of non-polynomial layers. Each curve corresponds to an individual non-polynomial layer. At left, we present the maximal and minimal values recorded at the input of the activation layers throughout the test set for each layer, where dashed lines denote activations in the MLP and solid lines represent attention activations. The middle and right graphs display, respectively, the mean and maximum values of the variance observed throughout the test set for each layer. The colors of the curves progress from blue to red to denote the sequence of layers: blue for the first layer, and red for the last, where intermediate layers are colored by interpolating blue and red based on their sequence position.}
    \label{fig:rangesMinmization}
\end{figure*}

{\noindent \textbf{Range Minimization} \quad}
 Our novel training method narrows the variance range at each \LayerNorm layer and the input to the activation layers, including both attention activations and MLP activations. To demonstrate the practicality and effectiveness of this method, we visualize in Fig.~\ref{fig:rangesMinmization} the maximal and mean-variance values at each layer, as well as the maximal and minimal values at the input of each activation. Both are measured at the end of each epoch to demonstrate progress during training. The graph clearly illustrates that prior to implementing the additional training phase, variance values were limited by 3300. This number was reduced to 300 during range training. Furthermore, it is evident that the activation range has been shortened from 70 to 20. These reductions greatly facilitate the approximation problem, since the error of the approximation increases with the domain width of the function being approximated, as reported in~\cite{baruch2023sensitive}. Additionally, this allows for the use of relatively low-degree polynomials, which significantly reduce the overall multiplication depth and, consequently, decrease the model's latency during secure inference. It is also noteworthy that the values of both the variance and activation norms tend to rise in the deeper layers, posing a greater challenge for approximating these layers.

{\noindent \textbf{Stabilized HE-Friendly ViT} \quad}
Although $\sigma$-transformers that normalize with \BatchNorm rather than \LayerNorm do not collapse and provide some non-trivial accuracy in the image classification regime, their performance still falls short of that of standard ViTs with \softmax attention and LayerNorm. To bridge this gap, Sections~\ref{subsec:heFrindelyNormalization} and \ref{subsec:HEFriendlyAttention} propose three techniques: (i) adding an additional \BatchNorm layer to the MLP in each ViT block (A), (ii) normalizing the attention matrix across the attention heads via \BatchNorm2D (QK), and (iii) implementing length scaling (S). In Table~\ref{tab:JustifyQK}, we ablate the contributions of each of these methods. These experiments were conducted using the Swin Transformer backbone on the Tiny ImageNet dataset. The results indicate that the incorporation of \BatchNorm instead \LayerNorm (denoted by B) initially decreases performance compared to the original \softmax-based Swin (denoted by O), with GELU-attention (G.) and ReLU-attention (R.) accuracy dropping to 39.0\% and 38.4\%, respectively. However, normalizing the attention matrix across the attention heads with \BatchNorm2D (B+QK) significantly improves accuracy. The subsequent addition of an extra \BatchNorm layer (B+QK+A) further enhances performance, nearly matching the original Swin's accuracy. Finally, the implementation of length scaling (B+QK+A+S) improves $\sigma$-attention-based models and closed the gap with the original Swin.

\begin{table}[h]
\centering
\small
\begin{tabular}{lccccc}\toprule
 $\sigma$ &O & B  & B+QK & B+QK+A & B+QK+A+S  \\\midrule
\midrule
 G.   &   59.4&   39.0 & 49.8 & 58.6  & 59.1 \\ 
 R.   &   59.4 &  38.4 & 50.1 & 56.2   & 58.6 \\\bottomrule
\end{tabular}
\caption{\textbf{Stabilize BatchNorm-based $\sigma$-ViT:} Experiments on both Tiny-ImageNet with $\sigma$-attention-based Swin transformer. We compare models with GELU-attention (G.) and \ReLU-attention (R.). `O' denotes the accuracy of the original Swin with \softmax and \LayerNorm. ``B' denote the accuracy of a $\sigma$-attention-based Swin with \BatchNorm. The remaining three columns represent the accuracy achieved by incorporating the first technique, the first two techniques combined, and all of the techniques, respectively.}
\label{tab:JustifyQK}
\end{table}

\subsection{Performance under HE\label{subsec:resultHE}}

We run HElayers \cite{helayers} version 1.52 as our HE SDK and set the underlying HE library to HEaaN. The concrete HE parameters were set as follows: We used ciphertexts with $2^{15}$ coefficients, a multiplication depth of 12, fractional part precision of 42, and integer part precision of 18. This context allows us to use up to 9 multiplications before bootstrapping is required. The security parameters were set to provide a solution with 128-bit security. Our hardware involved a computing system that used both CPU and GPU capabilities. The CPU component was an AMD EPYC 7763 64-core processor comprising 32 cores and 32 threads, along with 200 GB of RAM allocated to the processes under evaluation. Complementing this, we utilized an NVIDIA A100-SXM4-80GB GPU with 80GB of memory, which was integral for performing certain parts of the computation. This configuration was designed to exploit the combined processing power of CPU and GPU, ensuring efficient performance for our computational tasks. Under these settings, for a BERT-like transformer with 6 layers and 53.3M parameters (see hyperparameters in Tab.~\ref{tab:NLPhyperpams}, Appendix~\ref{sec:experimentalSetup}), secure inference over 128 tokens takes 305 seconds. For comparison, using the same hardware and SDK, secure inference with ResNet-152, which has 60M parameters, takes 432 seconds, setting the SOTA for CNNs over HE.

\section{Discussion and Future Work}
This paper presents an effective and innovative approach to converting transformers into a polynomial form via the scaled-$\sigma$ attention mechanism and a specialized training procedure that produces HE-friendly weights. Our techniques are the first to propose polynomial alternatives to the self-attention and \LayerNorm layers, allowing the deployment of secure inference with transformers for the first time. This advancement significantly extends the potential of the HE-based \gls{DL} models.

Looking ahead, we want to explore methods for directly approximating the self-attention and normalization layers by representing the \softmax function and the inverse square root with polynomial functions directly, without additional training stages. Moreover, we are interested in examining how our polynomial transformer variants compare to traditional models regarding vulnerability to adversarial attacks, performance on out-of-distribution samples, transfer learning capabilities, and interpretability. Lastly, we plan to scale our models for larger datasets, larger models, and more complex tasks, beyond the scope of classification, further broadening the horizons of HE-based \gls{DL}.

\newpage
\newpage
\newpage
\bibliography{custom}

\newpage
\appendix

\section{Our Polynomial Approximations\label{sec:appendixPolynomials}}
\begin{figure}[h]
    \centering
    \includegraphics[width=1.0\linewidth]{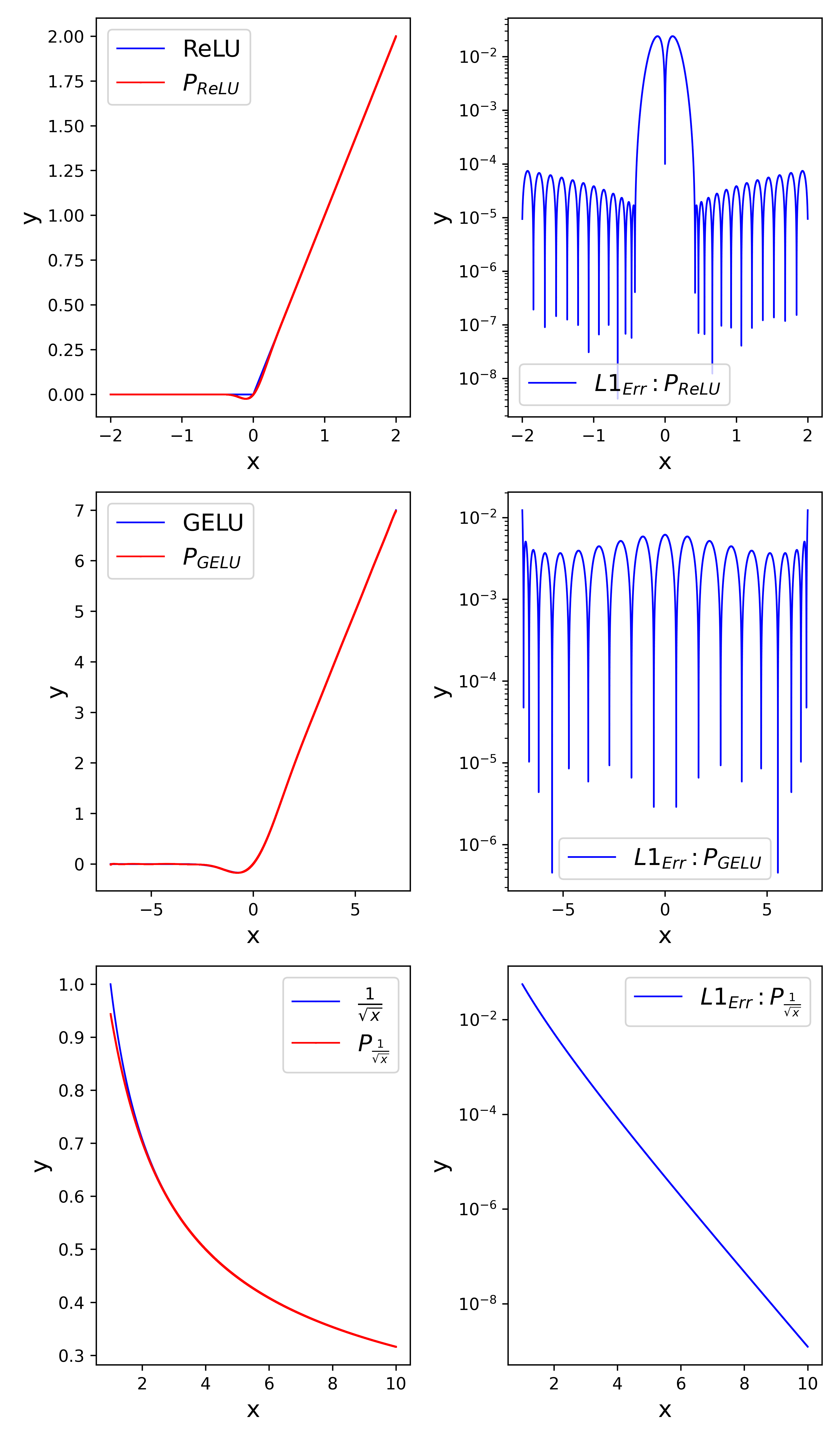}
    \caption{\textbf{Our Polynomial Approximations:}  (Left): Polynomial obtained by various well-known methods. (Right): L1 approximation error for each approximation presented on a logarithmic scale. At the top, we present polynomials and L1-error for ReLU; in the middle, for GELU, and at the bottom, for $\frac{1}{\sqrt{x}}$.}
    \label{fig:Polyaprox}
\end{figure}

After modifying the architecture and training using the methods described in Sec.~\ref{sec:method}, we obtain model architecture and weights that are HE-friendly. At this point, simple and well-known polynomial approximations can be applied to convert the model to a polynomial form. For the GELU and ReLU functions, which are used for attention activations and the MLP activations, we utilize the approximations from~\cite{lee2021precise} for ReLU, and polynomials generated with the Remez algorithm for GELU. To approximate the layer normalization, we use polynomials from~\cite{panda2022polynomial}. The polynomials and their approximation errors are presented in Fig. ~\ref{fig:Polyaprox}.

\section{Experimental Setup\label{sec:experimentalSetup}}
\paragraph{Training setup} All training experiments were performed on public datasets using a single A100 80GB GPU for a maximum of two days. All experiments were conducted using PyTorch, employing half-precision floating-point format. Results were averaged over three seeds, and all hyperparameters are detailed in Table~\ref{tab:NLPhyperpams} and Table~\ref{tab:Visionhyperpams}.
\begin{table}[h]
\centering
\small
\begin{tabular}{l c}
\toprule
\textbf{Parameter} & \textbf{Value} \\
\midrule
Model-width & 768 \\
Number of heads & 8 \\
$\sigma$-activation & ReLU \\
Context-length (training) & 512 \\
Batch-size & 256 \\
Optimizer & AdamW \\
Momentum & \( \beta_1, \beta_2 = 0.9, 0.999 \) \\
Base learning rate & $5e-4$ \\
Weight decay & 0.0 \\
Dropout & 0.1 \\
Training epochs & 100 \\
Learning rate schedule & cosine decay \\
Warmup epochs & 5 \\
Warmup schedule & linear \\ 
\bottomrule
\end{tabular}
\caption{Hyperparameters for Wikitext-103} 
\label{tab:NLPhyperpams}
\end{table}

\begin{table}[h]
\centering
\small
\caption{Hyperparameters for ViT and Swin} 
\label{tab:combinedhyperparams}
\begin{tabular}{l c c}
\toprule
\textbf{Parameter} & \textbf{ViT} & \textbf{Swin} \\
\midrule
Model-width & 192 & 96 \\
Model-depth & 9 & [2, 6, 4] \\
Number of heads & 12 & [3, 6, 12] \\
Trainable parameters & 2.7M & 7.13M \\
Window-size & - & 4 \\
\hline
Label smoothing & \multicolumn{2}{c}{0.1} \\
\(\sigma\)-activation & \multicolumn{2}{c}{GELU} \\
Patch-size & \multicolumn{2}{c}{\(4 \times 4\)} \\
Batch-size & \multicolumn{2}{c}{128} \\
Optimizer & \multicolumn{2}{c}{AdamW} \\
Momentum & \multicolumn{2}{c}{\( \beta_1, \beta_2 = 0.9, 0.999 \)} \\
Base learning rate & \multicolumn{2}{c}{\(1e-3\)} \\
Weight decay & \multicolumn{2}{c}{5e-2} \\
Dropout & \multicolumn{2}{c}{0.1} \\
Training epochs & \multicolumn{2}{c}{100} \\
Learning rate schedule & \multicolumn{2}{c}{cosine decay} \\
Warmup epochs & \multicolumn{2}{c}{10} \\
Warmup schedule & \multicolumn{2}{c}{linear} \\
Random erasing probability & \multicolumn{2}{c}{0.25} \\
\bottomrule
\end{tabular}
\label{tab:Visionhyperpams}
\end{table}
\end{document}